\UseRawInputEncoding
\documentclass[english, 10pt, twocolumn, twoside]{IEEEtran}
\usepackage{threeparttable}
\usepackage{mathrsfs}

\usepackage{cite}

\ifCLASSINFOpdf
\else
\fi
\usepackage{graphics}

\usepackage{subfigure}
\usepackage[cmex10]{amsmath}
\usepackage{amsfonts}
\usepackage{cases}
\usepackage{epsfig}
\usepackage{url}

\usepackage{algorithm}
\usepackage{algorithmic}

\usepackage{array}
\usepackage{amsmath}

\usepackage{multirow}
\usepackage{bigstrut}
\usepackage{hyperref}
\hypersetup{
	colorlinks=true
}
\usepackage{bm}
\usepackage{amsmath}
\usepackage{amssymb}
\usepackage{mathrsfs}
\usepackage{color}

\usepackage{booktabs}
\usepackage{multirow}
\usepackage{amsmath}
\usepackage[bottom]{footmisc}

\def\eg{\emph{e.g.}~}
\def\ie{\emph{i.e.},~}

\begin{document}

%
\title{Adversarial Dual-Student with Differentiable Spatial Warping for Semi-Supervised Semantic Segmentation}
%
%
%

\author{Cong~Cao,
Tianwei~Lin,
Dongliang~He,
Fu~Li,
Huanjing~Yue,~\IEEEmembership{Member,~IEEE,}\\
Jingyu~Yang,~\IEEEmembership{Senior Member,~IEEE,}
and~Errui~Ding
\thanks{This work was supported in part by the National Natural Science Foundation of China under Grant 62072331 and Grant 62231018.}
\thanks{C. Cao, H. Yue (corresponding author), and J. Yang are with the School of Electrical and Information Engineering,
Tianjin University, China. T. Lin, D. He, F. Li, and E. Ding are with the Department of Computer Vision (VIS) Technology, Baidu Inc.
}
}

\maketitle


\begin{abstract}
A common challenge posed to robust semantic segmentation is the expensive data annotation cost.
Existing semi-supervised solutions show great potential for solving this problem. Their key idea is constructing consistency regularization with unsupervised data augmentation from unlabeled data for model training. The perturbations for unlabeled data enable the consistency training loss, which benefits semi-supervised semantic segmentation. However, these perturbations destroy image context and introduce unnatural boundaries, which is harmful for semantic segmentation.
Besides, the widely adopted semi-supervised learning framework, i.e. mean-teacher,  suffers performance limitation since the student model finally converges to the teacher model.
%
In this paper, first of all, we propose a context friendly differentiable geometric warping to conduct unsupervised data augmentation; secondly, a novel adversarial dual-student framework is proposed to improve the Mean-Teacher from the following two aspects: (1) dual student models are learned independently except for a stabilization constraint to encourage exploiting model diversities; (2) adversarial training scheme is applied to both students and the discriminators are resorted to distinguish reliable pseudo-label of unlabeled data for self-training. Effectiveness is validated via extensive experiments on PASCAL VOC2012 and Cityscapes. Our solution significantly improves the performance and state-of-the-art results are achieved on both datasets. Remarkably, compared with fully supervision, our solution achieves comparable mIoU of 73.4\% using only 12.5$\%$ annotated data on PASCAL VOC2012. Our codes and models
are available at \href{https://github.com/cao-cong/ADS-SemiSeg}{https://github.com/cao-cong/ADS-SemiSeg}.
\end{abstract}

\begin{IEEEkeywords}
Semi-Supervised Semantic Segmentation, Dual-Student, Differentiable Spatial Warping.
\end{IEEEkeywords}

%
\IEEEpeerreviewmaketitle

\section{Introduction}
%
%
%
%

Understanding images at pixel level is prevalent due to that it enables many practical applications such as scene parsing, environment sensing for auto driving or visual navigation.
Therefore, semantic segmentation is becoming a more and more important computer vision task, which aims at semantically labeling natural images on per-pixel basis.
Recently, owing to the great progress in deep learning, numerous CNN frameworks have been developed for this task \cite{chen2014semantic, ronneberger2015u, chen2017deeplab, chen2017rethinking,zhao2017pyramid,yuan2018ocnet,sun2019high,ji2020encoder,athanasiadis2007semantic,meng2019weakly}.
However, the data driven training paradigm of CNNs requires a huge amount of manually labeled data. Especially for semantic segmentation, the annotations must be accurate at pixel level, leading to expensive cost for annotating an image. Therefore, data annotation poses great challenge for us to further improve the segmentation performance efficiently.

\begin{figure}[t]
  \centering
  \includegraphics[width=\columnwidth]{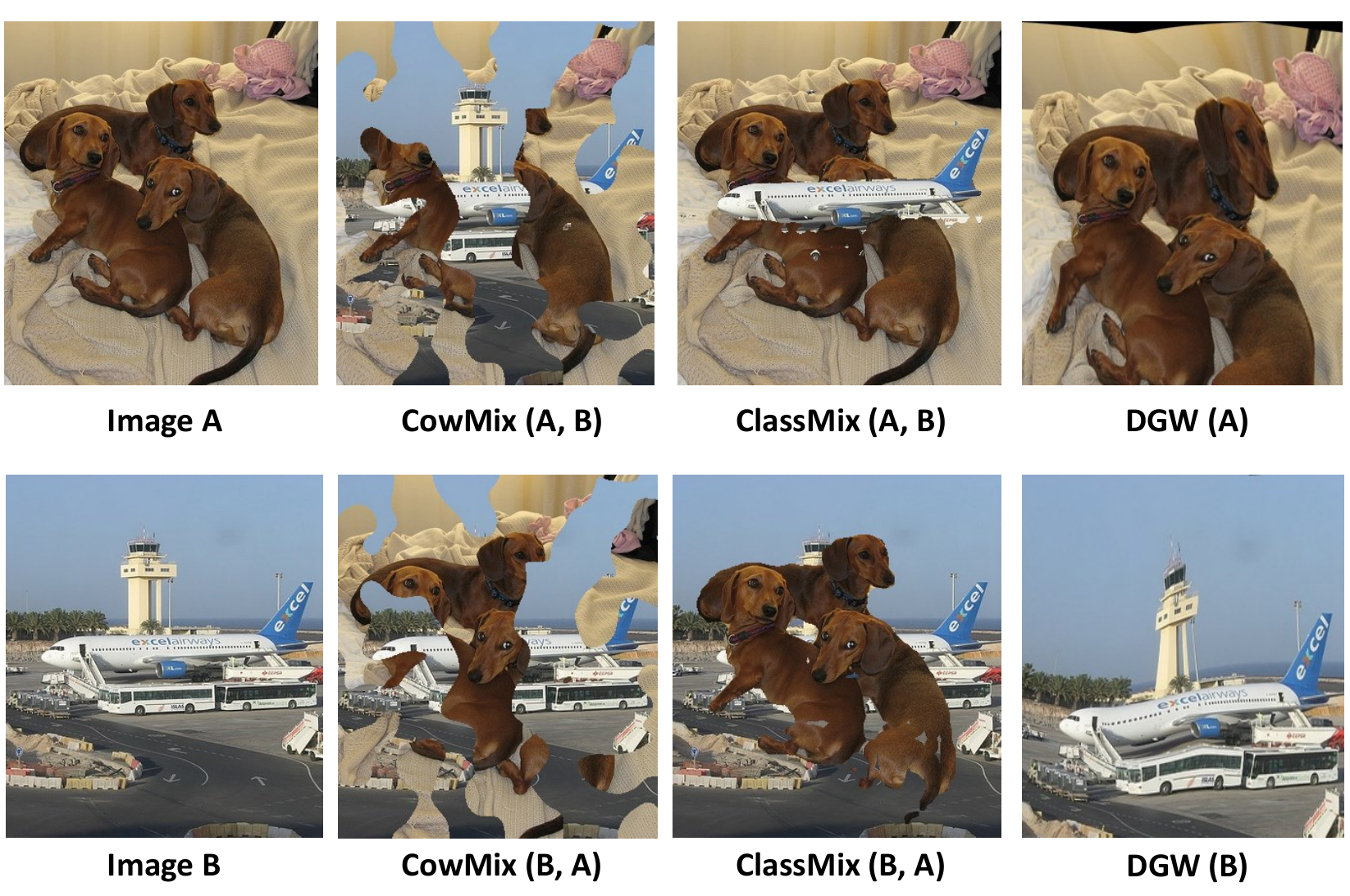}
   \caption{Illustration of heavily destroyed semantic context caused by CowMix and ClassMix. Our proposed differentiable geometric warping  (DGW) can eliminate this drawback.}
\label{illustration}
\end{figure}

Reviewing the literature, semi-supervised learning, which attempts to leverage unlabeled data to help model training, is widely utilized  to relieve annotation burden by the computer vision community. Many frameworks have been specially designed for semi-supervised semantic segmentation \cite{yun2019cutmix, french2019semi, olsson2020classmix}.
 A key trend of previous work is to apply unsupervised data augmentations, such as CutMix \cite{yun2019cutmix} and CowMix \cite{french2019semi}, to images and then construct consistency constraint. Although these strategies are able to improve the semi-supervised segmentation performance, they destroy the image context and introduce weird boundaries. The recently proposed ClassMix \cite{olsson2020classmix} achieves better performance since it replaces the weird boundaries by the class boundaries, as shown in Fig. \ref{illustration}. However, it still causes severe scene context destruction, harming semantic segmentation models.

%
%
The other widely utilized practice is to leverage the well-known state-of-the-art semi-supervised learning framework called mean-teacher \cite{tarvainen2017mean} and adapt it to semantic segmentation scenario for training under the consistency regularization.  However, mean-teacher ends up with a student model which is an exponential moving average version of the teacher model. Such moving average will make the student finally converge to the teacher model and therefore model diversity is not fully exploited. Under the mean-teacher framework, the performance of student model will be limited by the teacher \cite{ke2019dual}.

\textbf{To relieve the scene context destruction}, we propose a Differentiable Geometric Warping (DGW) mechanism to perform image perturbation. As can be seen from Fig. \ref{illustration}, our DGW distorts image content spatially and can better preserve semantic context compared to recent state-of-the-art CowMix \cite{french2019semi} and ClassMix \cite{olsson2020classmix}. For both rigid and non-rigid objects, CowMix and ClassMix will result in disharmonious surroundings around the object contours, \eg dogs over the sky and airplane near dogs. Context is proven to be of great importance for segmentation task \cite{chen2017deeplab}. Nevertheless, such discordant context caused by CowMix or ClassMix is not the case for real natural images, so it can be harmful for segmentation models. With our DGW, though the shape of rigid and non-rigid objects will be slightly distorted, it is still recognizable. Besides, the context near object contours is still natural. Therefore, for a segmentation model, correctly assigning class labels and precisely determining object boundaries for spatially geometric warped images are still possible. We also empirically verify that DGW achieves better performance than several commonly used image perturbations. In addition, it is worthy of noting that forcing models to predict consistency segmentation maps with respect to known spatial transformations is essential for stabilized model prediction.
Under the DGW setting, the extra supervision signal then becomes that, the prediction of distorted image should be consistent with the corresponding distorted prediction of the original image.
Specifically, in this paper, DGW is implemented based on Thin Plate Spline (TPS) \cite{bookstein1989principal} algorithm, which is a differentiable operation and fits a mapping from source to destination image coordinates.  

\textbf{To alleviate the drawback of mean-teacher}, in this paper, we propose to reform the recent state-of-the-art semi-supervised image classification framework named dual-student \cite{ke2019dual} to an adversarial setting for semi-supervised semantic segmentation. Our adversarial dual-student framework can: (1) better utilize unlabeled image data and (2) enable to exploit model diversity.  In more detail, two student networks are designed to learn from both supervised signal provided by labeled data and consistency regularization constructed with unlabeled data. The two students are relatively independent from each other and only reliable knowledge of both students can be exchanged. In this way, diversity (or variation) can be introduced and neither of the student model would be limited by the other one. Besides, adversarial training scheme is used for each student. For both student models, discriminators are trained to distinguish reliable predictions for unlabeled images and such reliable predictions are then leveraged as self-training signal to further improve our model.


In a nutshell, we focus on pushing the limitation of semi-supervised semantic segmentation  and our contributions can be summarized as follows:
\begin{itemize}
    \item We propose a differentiable geometric warping mechanism, which can be easily implemented with TPS and is context friendly, as unsupervised image augmentation for semi-supervised semantic segmentation.
    \item We are the first to reform dual-student framework to semantic segmentation task, and further propose the Adversarial Dual-Student (ADS) to benefit semi-supervised training from self-training mechanism.
    \item Extensive experiments show that our solution significantly improve the performance and new state-of-the-art performance is achieved on both PASCAL VOC2012 and Cityscapes datasets. Remarkably, compared with fully supervision, our solution achieves comparable mIoU of 73.4\% using only 12.5$\%$ annotated data on PASCAL VOC2012.
\end{itemize}

\section{Related Works}

\subsection{Semi-supervised Learning}
Semi-supervised learning is a classic topic in both computer vision and machine learning industry, which is recently drawing much attention in image classification task. In a major line, consistency regularization is used to take unlabeled data into account for defining the decision boundaries, thus can exploit unlabeled data to improve results \cite{laine2016temporal,tarvainen2017mean,miyato2018virtual,ji2019invariant,bachman2019learning}. Typically, mean-teacher \cite{tarvainen2017mean} proposes a teacher-student framework, which applies consistency-regularization between a student and a teacher based on Exponential Moving Average (EMA) of students in each update step.
Recently, Dual-Student \cite{ke2019dual} train two independent student networks and apply stabilization constraint instead of consistency constraint, avoiding performance bottleneck caused by coupled EMA teacher.
Another way is to predict for unlabeled data with a network and use the predictions as pseudo labels \cite{berthelot2019mixmatch,lee2013pseudo,tarvainen2017mean}.

\subsection{Semi-supervised Semantic Segmentation}
Compared with image classification, semi-supervised semantic segmentation is far more difficult and complex and also has two main lines.
One line exploits consistency regularization, \cite{yun2019cutmix,french2019semi,olsson2020classmix} focuses on different perturbation strategies, \cite{ouali2020semi,chen2021semi,liu2022perturbed} focuses on the architecture of consistency training.
The other line aims to encourage network to make confident predictions on unlabeled data. The work in \cite{souly2017semi,hung2018adversarial} uses an additional discriminator to select real enough predictions as pseudo labels. The work in \cite{feng2020semi,yang2022st++,fan2022ucc} uses dynamic self-training strategy to improve quality of pseudo labels. The work in \cite{mendel2020semi} uses an correction network to correct segmentation network’s prediction. The works in \cite{zhang2020robust} and \cite{zhang2020robust1} modify the segmentation network with self-attention module and apply spectral normalization to the GAN discriminator to improve the stability of the semi-supervised training. The work in \cite{chen2020digging} proposes a two-branch network to encode strong and pseudo label spaces respectively, which could well utilize pseudo-labels. Besides learning from pseudo labels, \cite{alonso2021semi} and \cite{zhou2021c3} utilize pixel-level contrastive learning to further improve the segmentation performance.
In addition, \cite{papandreou2015weakly, li2018weakly, lee2019ficklenet, ibrahim2020semi, ouali2020semi} combines semi-supervised and weakly-supervised semantic segmentation, \cite{kalluri2019universal, ouali2020semi} combines semi-supervised semantic segmentation and domain adaption. And there are some works \cite{sedai2017semi, chartsias2018factorised, cui2019semi, li2020self, hang2020local} designed for semi-supervised medical image segmentation.
Recently, \cite{mittal2019semi} combines the two lines via adding a multi-label mean teacher branch into the model proposed in \cite{hung2018adversarial} to further extent the mean teacher framework.
In this work, we combine the two lines via proposing an Adversarial Dual-Student framework. We design a segmentation stabilization consistency to obtain robust consistency, and introduce dual-discriminator to implicitly exploit unlabeled data.

\subsection{Perturbation Strategies}
In consistency regularization, predictions of unlabeled data are required to be invariant to perturbations. Recently, a lot of data augmentation methods \cite{zhang2017mixup, cubuk2018autoaugment, ict, zhao2019data, yun2019cutmix, french2019semi, lee2020learning, wang2020regularizing, olsson2020classmix} were proposed and some of them were applied to produce perturbations.
Among them, CutMix \cite{yun2019cutmix}, CowMix \cite{french2019semi} and ClassMix \cite{olsson2020classmix} are mask-based mixing method which cut out regions from one image and paste onto another.
Recently, ClassMix \cite{olsson2020classmix} combines images using binary masks generated upon  segmentation results instead of random rectangles and achieves state-of-the-art performance.
However, this kind of mixing usually leads to context distortion, generating unreasonable training pairs.  Therefore, instead of mask-based mixing, in this work, we suggest that segmentation consistency can be treated as a equivalent constraint problem \cite{jakab2018unsupervised,zhang2018unsupervised}, which requires that prediction is invariant to spatial warping and thus leads to robust prediction. To achieve this, we propose a differentiable spatial warping layer based on TPS.

\begin{figure*}[t]
\begin{center}
\begin{minipage}[b]{0.8\linewidth}
  \centering
  \centerline{\includegraphics[width=\linewidth]{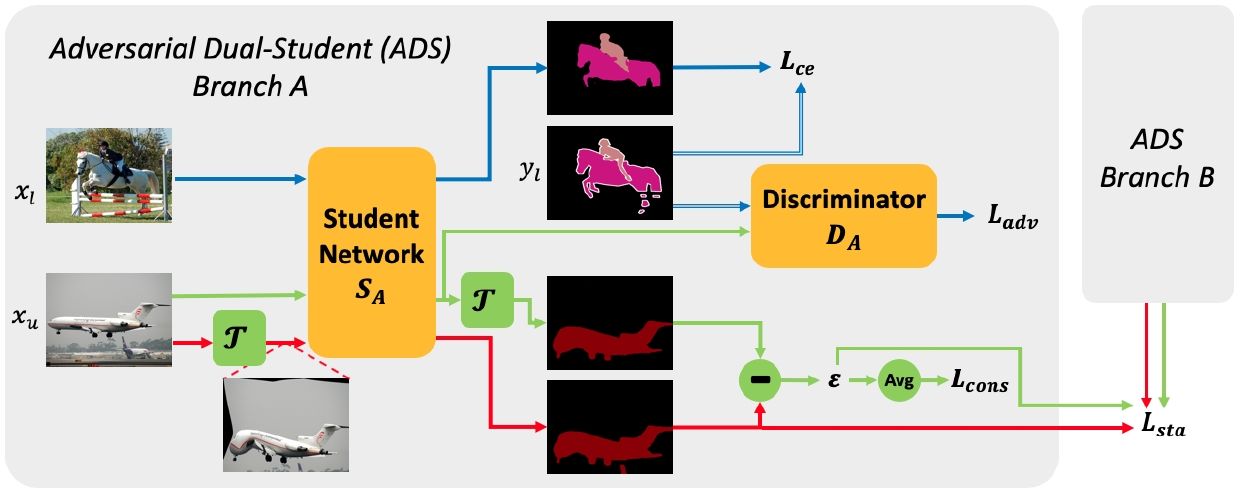}}
  \medskip
\end{minipage}
\end{center}
   \caption{Pipeline of our proposed ADS framework, where two branches have exactly the same architecture. Within each branch, supervised, consistency and adversarial learning loss functions are simultaneously adopted. DGW (denoted by $\mathcal{T}$) is used to disturb input and output geometrically. Between two branches, stabilization loss helps them to exchange reliable knowledge.}
\label{fig:framework}
\end{figure*}

\section{Approach}


\subsection{Overview}
\subsubsection{Problem Formulation}
Given an image $x \in R^{H\times W\times 3}$, it can be annotated with a semantic segmentation map $y \in R^{H\times W}$, where $y_{i,j} \in \left \{ 0, 1, ..., c  \right \}$ is the semantic label of pixel $(i,j)$ and $c$ is the number of target categories. In semi-supervised learning, for a full image set $X = \left \{ X_l, X_u \right \}$, annotations are only provided in labeled subset $X_l$, unlabeled subset is denoted by $X_u$. Our target is to train a semantic segmentation model with images from both labeled and unlabeled subset.

\subsubsection{Overall Framework}
The overall proposed framework is illustrated in  Fig. \ref{fig:framework}.
The ADS framework contains two exactly the same student branches without weight sharing.
Within each branch, labeled data contributes to supervised training loss, while the unlabeled data is used by consistency constraints. Both $x_u$ and its warped version $\mathcal{T}(x_u)$ are predicted via a student $S$ and we require that $\mathcal{T}(S(x_u))$ and $S(\mathcal{T}(x_u))$ should be well aligned. Adversarial learning is also equipped for both students. Adversarial training can result in a discriminator to distinguish high-quality pseudo labels for unlabeled data and such pseudo labels can be leveraged to improve the model training.
Between two branches, stabilization loss helps them exchange reliable knowledge.

\subsection{Differentiable Geometric Warping}
In the wild, objects may have various locations, poses and orientations, bringing great challenge to semantic segmentation.
On the other hand, rich potential context also exists in these natural image deformations.
For example, in unsupervised keypoints detection methods \cite{jakab2018unsupervised, zhang2018unsupervised}, forcing model to predict consistency keypoints with respect to known geometric transformations is the key factor for discovering stable keypoints.
Thus, how can we exploit these potential context in a controlled way to improve semantic segmentation?
We believe the semantic segmentation of objects should also be invariant to controlled geometric transformation, which can also lead to more stable segmentation results. Besides, the geometric transformation should meet the requirements of (1) objects in perturbed images should be recognizable for correct label assigning and (2) the contextual information near object contours should be natural for precise boundary determination.
To achieve this, we adopt TPS \cite{bookstein1989principal} deformation to construct our consistency regularization mechanism, which has following advantages: (1) can achieve more complex deformation than basic affine transform and be similar to natural image deformations, so the above requirements can be satisfied; (2) be differentiable thus can be easily combined with existing segmentation models during training; (3) source, and destination locations of transformation can be well controlled.

\begin{figure}[t]
\begin{center}
\begin{minipage}[b]{0.85\linewidth}
  \centering
  \centerline{\includegraphics[width=8.5cm]{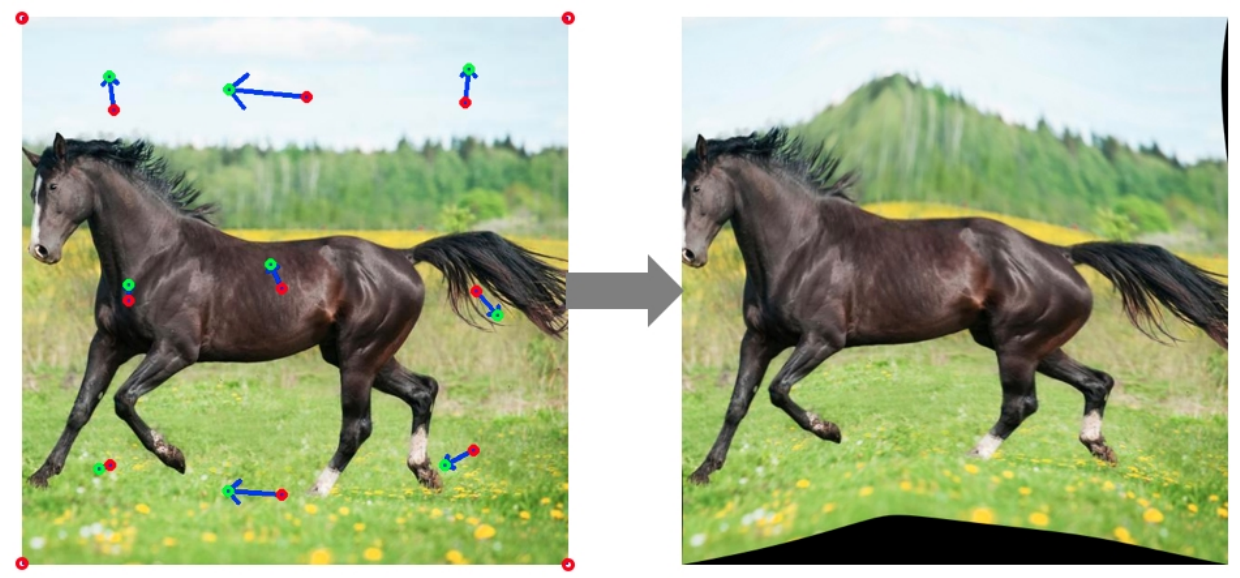}}
  \medskip
\end{minipage}
\end{center}
   \caption{Illustration of our proposed Differential Geometric Warping Layer, where warping is controlled by nine random source (red)-destination (green) point-pairs and four fixed ones.}
\label{fig:tps}
\end{figure}

Based on TPS, we propose the DGW
layer as a new consistency regularization mechanism. Fig. \ref{fig:tps} illustrates how DGW works. 
For an image $x \in R^{h\times w\times 3}$, we equally split it into $n\times n$ blocks.
Then, for the $i$-th block, we can obtain a source point as $p^s_i = ( u_i + \delta^s_u, v_i + \delta^s_v )$, where $(u_i, v_i)$ is the center point of block and $\delta^s \sim N(0, \sigma_s)$ is a  Gaussian function.
Further, we calculate corresponding destination point $p^d_i = ( u_i + \delta^s_u + \delta^d_u, v_i + \delta^s_v + \delta^d_v)$, where $\delta^d \sim N(0, \sigma_d)$.
Thus, we can obtain source point set $P_S = \{p^s_i\}_{i=1}^{N}$ and destination point set $P_D = \{p^d_i\}_{i=1}^{N}$, where $N = n\times n + 4$ and here we include 4 fixed corner points in the source and destination point sets. The corner points are used for stabilizing transformation. Finally, $P_S$ and $P_D$ are used to control the TPS transformation to generate warped image $\mathcal{T}(x)$, where the deformation degree is controlled by $\sigma_d$.

\subsection{Adversarial Dual-Student Framework}
In this section, we will introduce the details of our proposed
ADS framework. First, we introduce how we reform the Dual-Student framework for semantic segmentation task based on DGW. Then, the details of adversarial training are introduced. Fig. \ref{fig:framework} presents the schematic diagram of our ADS framework.

\subsubsection{Dual-Student}
Most existing semi-supervised semantic segmentation methods \cite{mittal2019semi, yun2019cutmix} adopt mean-teacher framework. However, as discussed in \cite{ke2019dual}, because of exponential moving average, teacher network in mean-teacher tends to be very close to the student when the training process converges, causing the lacking of diversity and performance bottleneck.
To overcome this coupled teacher problem, Dual-Student \cite{ke2019dual} is proposed to use two independently initialized student network without teacher.
%
The core idea of Dual-Student is applying a stabilization constraint to unlabeled data to exchange reliable knowledge.
In this work, we propose a novel consistency constraint and  a novel stabilization constraint for semantic segmentation based on Dual-Student. We denote the two student segmentation networks as $S_a$ and $S_b$ respectively.
In the following, we introduce our ADS framework via introducing each loss function. For simplicity, we only show loss functions of $S_a$. The losses for the other student model $S_b$ can be analogically calculated.

\subsubsection{Cross-Entropy Loss}
For labeled samples $X_l$, we directly apply supervised pixel-level cross entropy loss function $L_{ce}$ for each network.
With the network output $S_a(x) \in R^{H\times W\times C}$, we can calculate cross-entropy loss as:

\begin{equation}
L_{ce}^a = -  \sum_{i,j,c}{y_{l_{(i,j,c)}}logS_a(x_l)_{(i,j,c)}}
\end{equation}

\subsubsection{Consistency Constraint}
To exploit context of unlabeled samples $X_u$, we apply consistency constraint based on DGW transform for each branch separately.
For an unlabeled image $x_u$, segmentation network $S_a$ is applied twice, where DGW transform with the same control points is applied before and after $S$ separately.
%
With two generated results $S_a(\mathcal{T}(x_u))$ and $\mathcal{T}(S_a(x_u))$, we can calculate a pixel-wise consistency error $\varepsilon^a \in R^{h\times w}$, which is used for calculating both consistency and stabilization loss:

\begin{equation}
\varepsilon^a(x_u)_{(i,j)} = \left(S_a(\mathcal{T}(x_u))_{(i,j)} - \mathcal{T}(S_a(x_u))_{(i,j)}\right)^2
\end{equation}
where $\mathcal{T}$ is DGW transform used for spatial perturbation. Then, based on  $\varepsilon^a$, we can obtain consistency loss as:
\begin{equation}
L_{cons}^a =  \frac{1}{h\times w}\sum_{i,j}^{h,w}\varepsilon^a(x_u)_{(i,j)}.
\end{equation}
This consistency constraint encourages each student network to be invariant to geometric transform, leading network to learn how to handle natural objects with complex shapes, orientations, and deformations.

\subsubsection{Stabilization Constraint}
The core idea of Dual-Student \cite{ke2019dual} is that only reliable knowledge could be exchanged between two student networks.
To achieve this, Dual-Student first defines stable samples, and then elaborates the derived stabilization constraint to exchange knowledge.
In this work, we further extend the concept of stable samples and stabilization constraint from image-level to pixel-level to achieve semi-supervised semantic segmentation.


What is a stable pixel in semantic segmentation? We suggest that a stable pixel should be invariant to spatial perturbation. Thus, for each student network $S$, a stable pixel should meet following requirements: (1) \textbf{semantically the same}: the predicted semantic label of $S_a(\mathcal{T}(x))_{(i,j)}$ and $\mathcal{T}(S_a(x))_{(i,j)}$ should be the same; (2) \textbf{high confidence}:  either of $S_a(\mathcal{T}(x))_{(i,j)}$ and $\mathcal{T}(S_a(x))_{(i,j)}$  should be larger than a pre-defined threshold $\xi$. If a pixel $(i,j)$ satisfies these two conditions at student A, we denote stable score $r^a_{(i,j)} = 1$, otherwise 0.
Then, we can derive the stabilization constraint in different situations:
(1) if a pixel is not stable,  no constraint will be applied;
(2) if a pixel is stable for one student but unstable for the other, the stable one is used to constrain the unstable one.
(3) if a pixel is stable for both student, then we use the more stable one to constrain the other one. 
Thus, for student A, we can calculate pixel-wise stabilization loss as:
\begin{equation}
l^a_{sta}(x) = \left\{\begin{matrix}
[\varepsilon^a(x)<\varepsilon^b(x)]_1L_{mse}(x) & r^a=r^b=1 \\
r^aL_{mse}(x)  & otherwise,
\end{matrix}\right.
\label{eeee}
\end{equation}
where $[ ]_1$ is a indicator and it equals 1 only when student A is more stable, \ie $\varepsilon^a(x)<\varepsilon^b(x)$. Otherwise, it is 0.
$L_{mse}$ measures the mean square error between outputs of two student networks:
\begin{equation}
L_{mse}(x_u)_{(i,j)} = (S_a(\mathcal{T}(x_u))_{(i,j)} - S_b(\mathcal{T}(x_u))_{(i,j)})^2,
\label{eeeee}
\end{equation}
For simplicity, we omit the pixel position $(i,j)$ in Eq. \ref{eeee}. 
Finally, the total stabilization loss can be calculated as
\begin{equation}
L_{sta}^a(x_u) = \frac{1}{h\times w}\sum_{i,j}^{h,w} {l^a_{sta}(x_u)_{(i,j)}}
\end{equation}
The proposed stabilization loss can effectively exchange reliable knowledge between two segmentation networks, meanwhile keeping the diversity.

\subsubsection{Adversarial learning}
To further exploit knowledge in unlabeled data, we combine conditional adversarial learning with our dual-student framework. Specifically, we adopt the discriminator of s4GAN \cite{mittal2019semi} as the adversarial module in each branch of Dual-Student.
Formally, an extra discriminator $D_a$ needs to be trained for $S_a$ with original GAN loss as:

\begin{equation}
L_D^a = \mathbb{E}[logD_a(y_l \oplus x_l)] + \mathbb{E}[log(1-D_a(S_a(x_u)\oplus x_u))],
\end{equation}
where $\oplus$ denotes concatenation along channels.
Based on discriminator, we can calculate feature matching loss to minimize the mean discrepancy of the feature statistics between $S_a(x_u)$ and $y_l$:

\begin{equation}
L_{fm}^a = ||\mathbb{E}[D_a^{(k)}(y_l\oplus x_l)] - \mathbb{E}[D_a^{(k)}(S_a(x_u)\oplus x_u)]  ||,
\end{equation}
where $D_a^{(k)}$ denotes the $k$-th intermediate feature map of $D_a$.
Furthermore, following \cite{mittal2019semi}, we adopt self-training loss $L_{st}^a$ to pick the well predicted pixels (which can fool $D_a$ and are predicted as ``Real'' with score larger than some threshold) for supervising $S_a$.
The loss term $L_{st}^a$ is defined as:
\begin{equation}
    L_{st}^a = \left\{\begin{matrix}
- \Sigma_{h,w,c} \  \hat{y}\text{log}S_a(x^u), \text{if}~ D_a(S_a(x^u)\oplus x^u) \ge \gamma, \\
0, \text{otherwise},
\end{matrix}\right.
\end{equation}
where $\hat{y}$ denotes pseudo pixel-wise labels generated from segmentation result $S_a(x^u)$ and $\gamma$ is a threshold that determines how confident about the prediction that $D_a$ needs to be. In our implementation, $\gamma$ is set to be 0.6 and 0.9 for PASCAL VOC2012 and Cityscapes, respectively. $\hat{y}$ is generated by sharpen the prediction value from continuous real values to be discrete value in $\{0,1\}$ by converting logit to one-hot pseudo-label.
We denote the overall adversarial loss as:
\begin{equation}
L_{adv}^a = \lambda_{fm} \cdot L_{fm}^a + L_D^a + \lambda_{st} \cdot L_{st}^a,
\end{equation}
where $\lambda_{fm}, \lambda_{st}$ are weighting factors.

\subsubsection{Training Objective}
With aforementioned loss functions, for student model $S_a$, the loss is combined as:
\begin{equation}
L^a = L_{ce}^a + \lambda_1 \cdot L_{cons}^a + \lambda_2 \cdot L_{sta}^a+\lambda_3 \cdot L_{adv}^a.
\end{equation}
The overall training objective is to minimize the following loss function $L=L^a+L^b$.


\section{Experiments}

\subsection{Datasets and Setup}
\textbf{Datasets.}
\emph{PASCAL VOC2012}  \cite{everingham2015pascal} dataset consists 20 foreground object classes and one background class. Following \cite{olsson2020classmix}, we use the original images along with the extra annotated images from the Segmentation Boundary Dataset (SBD) \cite{hariharan2011semantic}, totally including 10582 training and 1449 validation images.
\emph{Cityscapes} \cite{cordts2016cityscapes} is an urban driving scenery dataset. It has 2975, 500 and 1525 annotated images for training, validation and testing respectively, and there are 19 semantic categories in total.
\emph{PASCAL-Context} \cite{mottaghi2014role} is a whole scene parsing dataset containing 4,998 training and 5,105 testing images with dense semantic labels. Following \cite{mittal2019semi}, we utilize semantic labels for the most frequent 60 classes (including the background class).

For all our experiments, mean Intersection-over-Union (mIoU) is reported.

\textbf{Implementation Details.}
%
The training of our method is performed with one Nvidia V100 GPU.
%
 For convenient comparison and proof of concept purposes, extensive experimental results are provided with DeepLab-v2 \cite{chen2017deeplab}, which are also common practice in previous works \cite{hung2018adversarial, olsson2020classmix}. ResNet101 \cite{He_2016_CVPR} pretrained on MS-COCO \cite{lin2014microsoft} is used as our backbone. Since some state-of-the-art works utilize DeepLab-v3+ as baseline (where pretrained ResNet50 serves as the backbone), we also provide our results with DeepLab-v3+ as baseline.
%
For the student models, we adopt SGD optimizer with a base learning rate of $2.5\times 10^{-4}$ and the learning rate is decreased with polynomial decay of power 0.9. Momentum is set to 0.9 and weight decay is $5\times 10^{-4}$.
As for the two discriminators, Adam optimizer with a base learning rate of  $1\times 10^{-4}$ is used and learning rate is decreased with polynomial decay of power 0.9.
Following \cite{hung2018adversarial}, for PASCAL VOC2012, we augment data with random scaling, cropping, and horizontal flipping. Scaling factors are randomly chosen between 0.5 and 1.5, cropping size is $321\times 321$. For Cityscapes, images are resized to $256\times 512$ without random cropping and scaling. For PASCAL-Context, images are random cropped to $321\times 321$. Following previous practices in \cite{mittal2019semi}, hyperparamters of $\lambda_{fm}$, $\lambda_{st}$ are empirically set to 0.1, 1, and $\lambda_1$, $\lambda_2$ and $\lambda_3$ equal 10, 1 and 100, respectively. Our models are trained for 90000, 80000, and 80000 iterations on PASCAL VOC2012, Cityscapes, and PASCAL-Context, respectively. %
In inference phase, the student network with better performance is used.
The input image resolutions are set to 321$\times$321 and 256$\times$512 on PASCAL VOC2012 and Cityscapes, respectively.

\subsection{Ablation Study}

\subsubsection{Contribution of Each Component}
To demonstrate the effectiveness of the proposed DGW and ADS, we conduct an ablation study on PASCAL VOC2012 and Cityscapes. As shown in Table \ref{label1}, when DGW is used to construct a Mean-Teacher framework, the performance can be largely improved by about 3\%-5\% in terms of mIoU. When adversarial training is added, the performance is further improved by about 0.6\%-1.2\%. By replacing the Mean-Teacher scheme with Dual Student, the mIoU can be boosted from 70.6\% and 73.2\% to 71.5\% and 73.5\% for PASCAL VOC2012, from 60.1\% and 63.2\% to 61.1\% and 64.4\% for Cityscapes, when 1/8 and 1/4 of labeled data are used, respectively. It verifies that adopting Dual-Student is more robust than Mean-Teacher. From the last two rows, it can be concluded that stabilization loss also remarkably contributes to the final performance.

\begin{table}[t]
\centering
\caption{Evaluation on the effectiveness of each component on PASCAL VOC2012 and Cityscapes. We report mIoUs when only 1/8 and 1/4 training samples are labeled. MT, DS and Adv denote Mean-Teacher, Dual-Student and adversarial training ($L_{fm}+L_D+L_{st}$), respectively.}
\begin{tabular}{m{4cm}m{0.45cm}<{\centering}m{0.45cm}<{\centering}m{0.45cm}<{\centering}m{0.45cm}<{\centering}}
\toprule
Dataset & \multicolumn{2}{c}{VOC2012} & \multicolumn{2}{c}{Cityscapes} \\
\hline
Method  & 1/8 & 1/4 & 1/8 & 1/4 \\
\hline
Supervised DeepLab-v2            & 65.2 & 68.9 & 55.0 & 60.3 \\
\hline
MT + DGW                    & 70.6 & 73.2 & 60.1 & 63.2 \\
MT + DGW + Adv            & 71.8 & 74.2 & 60.8 & 63.8 \\
DS + DGW                    & 71.5 & 73.5 & 61.1 & 64.4 \\
DS + DGW + Adv - $L_{sta}$ & 72.7 & 73.7 & 61.0 & 64.2 \\
DS + DGW + Adv  & 73.4 & 74.9 & 62.3 & 66.2 \\
\bottomrule
\end{tabular}
\label{label1}
\end{table}

\begin{table}[t]
\centering
\caption{Evaluation on different perturbation methods and different configurations of DGW on PASCAL VOC2012. The experimental results are based on DeepLab-v2 backbone. In this experiment, our adversarial Dual-Student framework is used and we only modify the augmentation strategy $\mathcal{T}$.}
\begin{tabular}{m{5cm}m{0.7cm}<{\centering}m{0.7cm}<{\centering} }
\toprule
 Methods   & 1/8 & 1/4 \\
\hline
CutMix &69.1 &70.7 \\
CowMix    & 71.5 & 72.4 \\
ClassMix  & 72.5 & 74.1 \\
Basic Affine Transform &72.5 &74.3 \\
\hline
$Rand_{3\cdot3}$ & 72.3 & 73.9 \\
$Grid_{3\cdot3}$ ($Rand_D$) & 72.9 & 74.3 \\
$Grid_{3\cdot3}$ ($Rand_S$ + $Rand_D$, ours) & 73.4 & 74.9 \\
\hline
$Grid_{2\cdot2}$ ($Rand_S$ + $Rand_D$, ours) & 72.8 & - \\
$Grid_{4\cdot4}$ ($Rand_S$ + $Rand_D$, ours) & 74.2 & - \\
$Grid_{5\cdot5}$ ($Rand_S$ + $Rand_D$, ours) & 72.6 & - \\
\bottomrule
\end{tabular}
\label{label3}
\end{table}

\begin{figure*}[t]
\begin{center}
\begin{minipage}[b]{0.95\linewidth}
  \centering
  \centerline{\includegraphics[width=\linewidth]{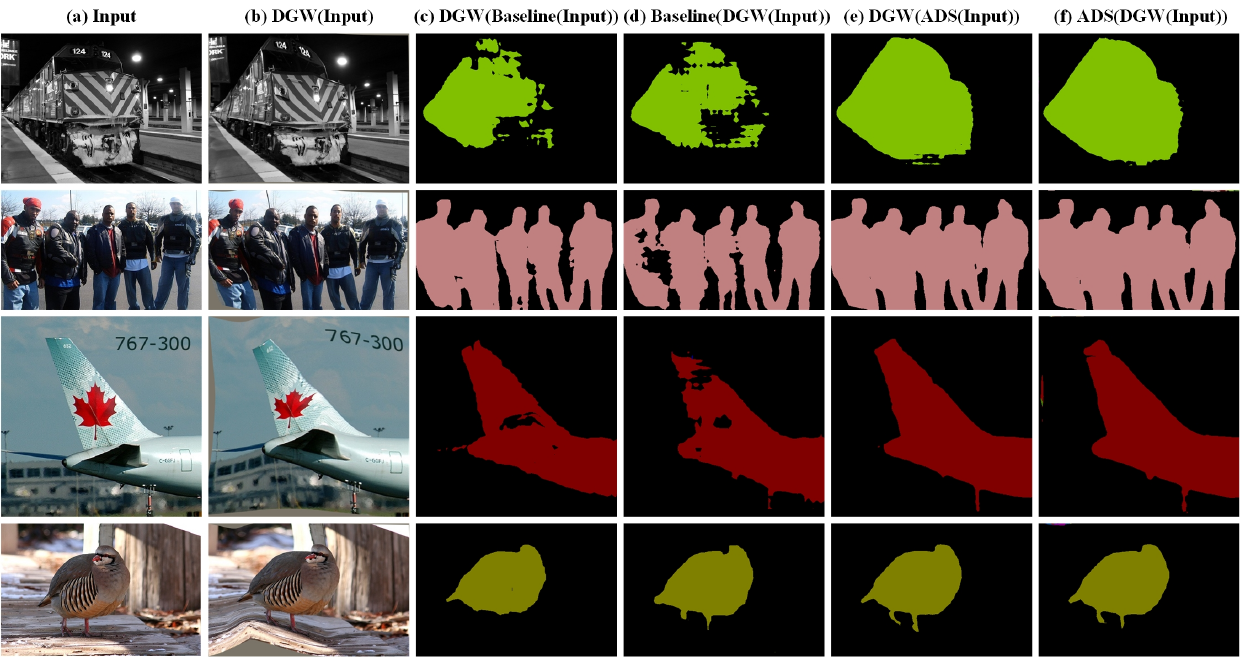}}
  \medskip
\end{minipage}
\end{center}
   \caption{Visual comparisons between DGW(Net(Input)) and Net(DGW(input)) when 1/8 labeled data is available, where Net is set to DeepLab-v2 (Baseline) and Ours (ADS), respectively. For each row, from left to right: (a) input image, (b) input image processed by DGW, (c) the predicted result generated by DGW(Baseline(Input)), (d) the predicted result generated by Baseline(DGW(input)), (e) the predicted result generated by DGW(ADS(Input)), and (f) the predicted result generated by ADS(DGW(input)).}
\label{fig:vs_consistency}
\end{figure*}

\begin{figure*}[t]
\begin{center}
\begin{minipage}[b]{0.95\linewidth}
  \centering
  \centerline{\includegraphics[width=\linewidth]{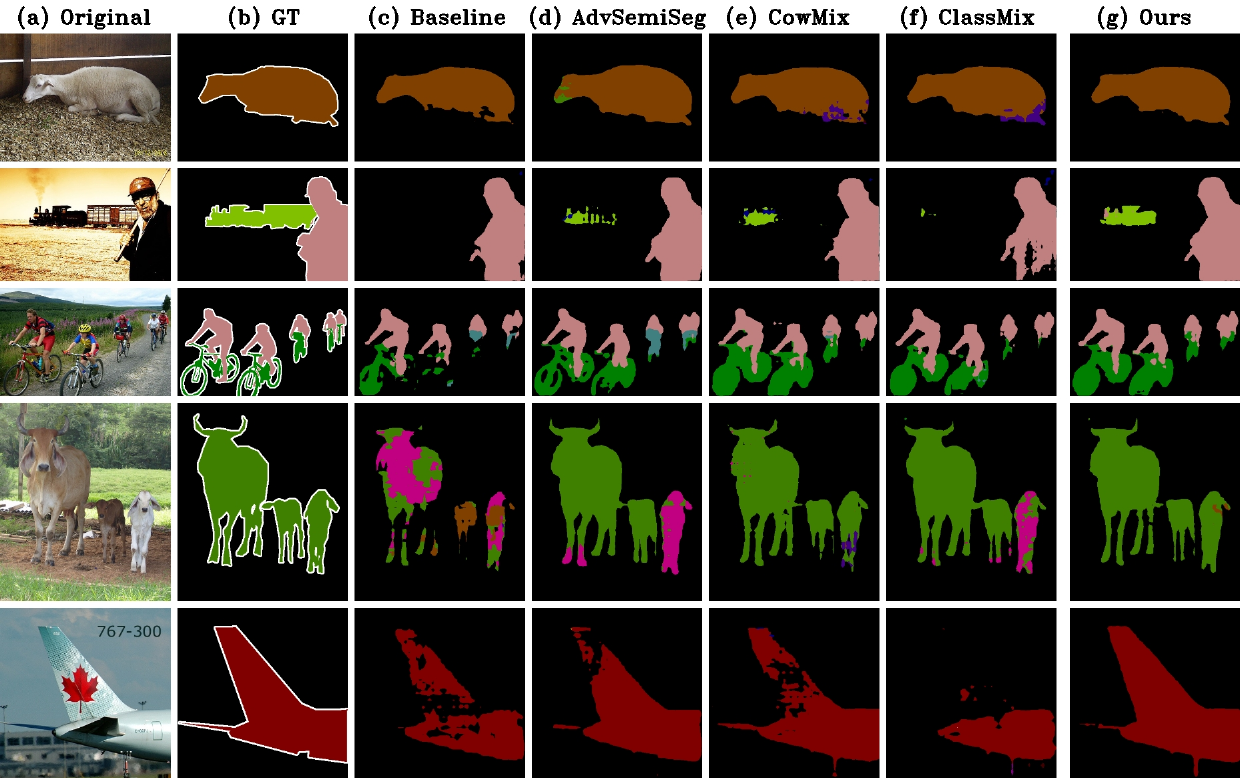}}
  \medskip
\end{minipage}
\end{center}
   \caption{Qualitative comparison results on PASCAL VOC2012 using 5$\%$ labeled samples. Our proposed semi-supervised approach produces improved results compared to the baseline, and also outperforms other state-of-the-art methods.}
\label{fig:vs_pascal}
\end{figure*}

\begin{figure*}[t]
\begin{center}
\begin{minipage}[b]{0.95\linewidth}
  \centering
  \centerline{\includegraphics[width=\linewidth]{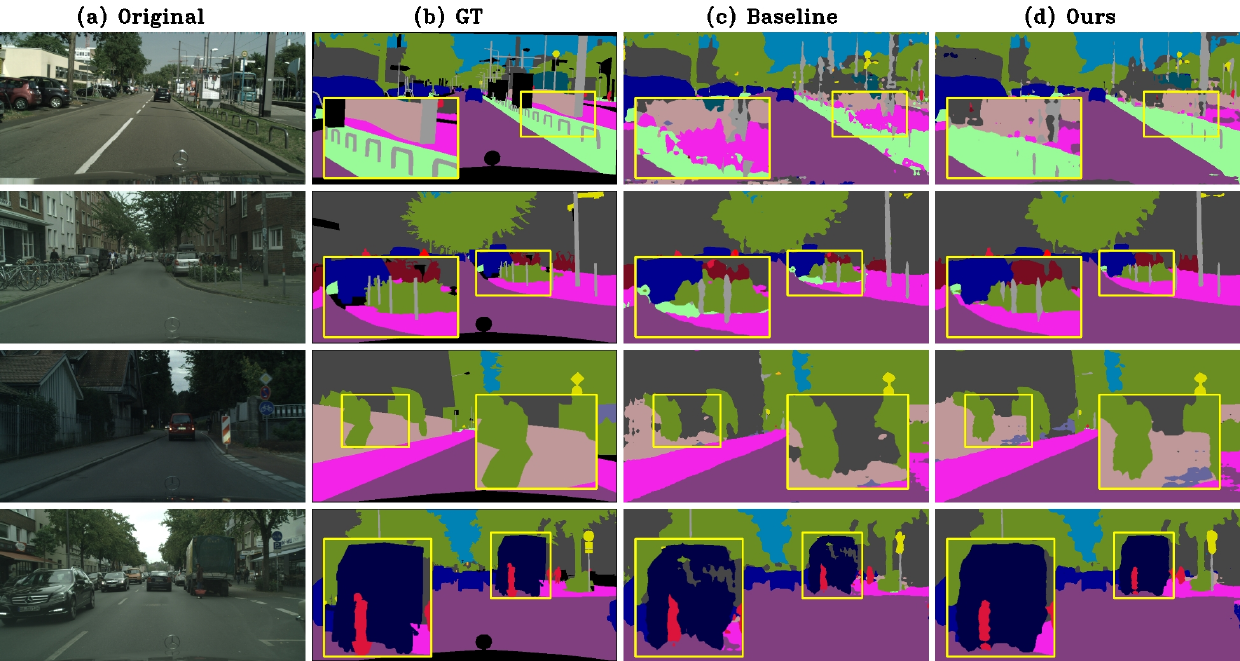}}
  \medskip
\end{minipage}
\end{center}
   \caption{Qualitative results on Cityscapes using 1/4 labeled samples. Our proposed method produces significantly improved results compared to the baseline. To illustrate the differences clearly, in each segmentation map, an area is enlarged.}
\label{fig:vs_city}
\end{figure*}

\subsubsection{DGW vs. Other Perturbation Methods}
Though ICT \cite{ict}, CutOut and CowOut \cite{french2019semi} made early attempts on unsupervised augmentation for semi-supervised semantic segmentation, we choose to compare DGW with several recent popular unsupervised data augmentation strategies, such as CutMix and CowMix proposed in \cite{yun2019cutmix} and ClassMix \cite{olsson2020classmix}, since they are verified to be better perturbations for the semi-supervised segmentation task. We also compare DGW with basic affine transform which is constructed by random translations, scalings, rotations, and shears. From the top four rows of Table \ref{label3}, we can see that when the overall framework is the same and we replace DGW with other augmentation mechanisms, the performance will be significantly degraded. Specifically, the performance of DGW not only outperforms the class-level augmentation methods, such as Class-Mix, Cut-Mix, and CowMix, but also outperforms the simple geometric-transform-based augmentation, i.e., basic affine transform. The main reason is that the proposed DGW can flexibly enrich the diversity of objects and is image context friendly.

\subsubsection{Configuration of DGW}
Finally, we evaluate how a specific configuration of DGW makes a difference to the performance. In Table \ref{label3}, $Rand_{3\cdot3}$ means the nine source and destination control points are randomly chosen, $Grid_{3\cdot3}$ means to choose the centers of 9 equally divided image grids,  $Rand_S$ and $Rand_D$ denote source and destination points are allowed to have a random Gaussian offsets to the grid centers, respectively. The results show that totally random sampling of control points is not a good choice. It is reasonable because it may result in sever distortion with certain probability. Grid sampling is much better, especially when both source and destination points are allowed to exhibit slight randomness. For block number n, we use n=3 as the default setting. In Table \ref{label3}, we verify the influence of n and give the corresponding results when n=2,3,4,5 respectively. The results show that n=3 leads to better results than n=2 or 5. Note that, n=4 leads to the best result, which demonstrates the potential of our DGW augmentation.

\subsubsection{Impact of Hyperparameters}
There are several hyperparameters in our framework. For the weighting $\lambda$'s, we do not tune with them and simply follow previous wisdom in \cite{ke2019dual,mittal2019semi}. Carefully tuning these parameters could bring more impressive results and we leave it as our future work. The settings of $\sigma_s$ and $\sigma_d$ are not quite sensitive to the choice of $\sigma_s$ and $\sigma_d$.  We provide empirical results on PASCAL VOC2012 (Table \ref{table-hyper}) for evidence. We adjust $\sigma_s$ and $\sigma_d$ from 0.05 to 0.15 and 0.15 to 0.25, respectively. Results on PASCAL VOC2012 with only 1/8 training data are shown in Table \ref{table-hyper}.
From these results, we can see that under a wide range of settings of $\sigma_s$ and $\sigma_d$, our ADS framework can achieve very good mIoU performance on PASCAL VOC2012. When only 1/8 training data is used, the performance is consistently around 73\% in terms of mIoU. It shows that our DGW is robust and beneficial. In our experiments, $\sigma_s$ and $\sigma_d$ are set to be 0.1 and 0.2, respectively.

\subsubsection{Effect of Consistency in Geometric Transformation}

To verify the effect of consistency in the geometric transformation, we calculate mIoU between DGW(Net(Input)) and Net(DGW(input)), and the experimental results are shown in Table \ref{table:consistency}. The higher mIoU means the output of DGW(Network(Input)) is closer to the output of Network(DGW(input)), and that is, the network has better geometric transformation consistency. From Table IV, we can see that our proposed method has better geometric transformation consistency than the baseline for each data condition due to our proposed consistency loss. Besides, the model trained on more labeled data not only has a better segmentation result but also has a better geometric transformation consistency than that trained on less labeled data. It verifies that a better model has better geometric transformation consistency, and the consistency loss based on geometric transformation could improve the performance of segmentation. The qualitative results are shown in Fig. \ref{fig:vs_consistency}, which also demonstrates that our proposed method leads to better geometric transformation consistency and better segmentation results.

\begin{table}[ht]
\caption{The mIoU of ADS under different settings of $\sigma_s$ and $\sigma_d$. The experimental results are based on DeepLab-v2 backbone.}
\begin{center}
\begin{tabular}{c|c|c|c|c}
	\hline
	\multicolumn{2}{c|}{\multirow{2}{*}{mIoU}} & \multicolumn{3}{c}{$\sigma_d$} \\ \cline{3-5}
	\multicolumn{2}{c|}{}&0.15&0.2 &0.25\\ \hline
	\multicolumn{1}{c|}{\multirow{3}{*}{$\sigma_s$}} &0.05 &72.8 &73.3 &73.0 \\
	\cline{2-5}
	&0.1 &73.0 &73.4 &73.4 \\
	\cline{2-5}
	&0.15 &73.0 &73.3 &73.1 \\
	\hline
\end{tabular}
\end{center}
\label{table-hyper}
\end{table}

\begin{table}[h]
\centering
\caption{Comparison of consistency (in terms of mIoU) between DGW(Net(Input)) and Net(DGW(input)), where Net is set to DeepLab-v2 (the first row) and Ours (the second row).}
\begin{tabular}{m{2.8cm}<{\centering}m{0.5cm}<{\centering}m{0.5cm}<{\centering}m{0.5cm}<{\centering}m{0.5cm}<{\centering}m{0.45cm}<{\centering}}
\toprule
Methods &1/100 &1/50 &1/20 &1/8 & 1/4 \\
\hline
DeepLab-v2\cite{chen2017deeplab}  & 57.2 & 68.0 & 72.4 & 73.8 & 74.8 \\
\hline
Ours        & 75.9 & 81.4 & 84.0 & 84.5 & 84.6 \\
\bottomrule
\end{tabular}
\label{table:consistency}
\end{table}

\subsection{Comparison with State-of-the-Arts}


\begin{table}[h]
\centering
\caption{Comparison between our proposed method with SOTA semi-supervised methods on PASCAL VOC2012, where DeepLab-V2 is a supervised baseline. \emph{Rela. Impr.} here stands for relative improvement over baseline. $^*$: Its backbone is DeepLab-v2, however, it uses a DeepLab-v3 model as its corrector to help training its backbone.}
\begin{tabular}{m{2.8cm}<{\centering}m{0.5cm}<{\centering}m{0.5cm}<{\centering}m{0.5cm}<{\centering}m{0.5cm}<{\centering}m{0.45cm}<{\centering}m{0.45cm}<{\centering}}
\toprule
Methods &1/100 &1/50 &1/20 &1/8 & 1/4 &full\\
\hline
DeepLab-v2\cite{chen2017deeplab}  & 42.6 & 53.1 & 60.1 & 65.2 & 68.9 & 73.4 \\
\hline
AdvSemiSeg\cite{hung2018adversarial}  & 38.8 & 57.2 & 64.7 & 69.5 & 72.1 & 74.9 \\
CowMix\cite{french2019semi}     & 52.1 & -    & 67.2 & 69.3 & 71.2 & -  \\
MLMT+S4GAN\cite{mittal2019semi}  & -    & 63.3 & 67.2 & 71.4 & -    & 75.6 \\
DST-CBC\cite{feng2020semi}     & \textbf{61.6} & 65.5 & 69.3 & 70.7 & 71.8 & 73.5 \\
ClassMix\cite{olsson2020classmix}    & 54.2 & 66.2 & 67.8 & 71.0 & 72.5 & - \\
ECS$^*$\cite{mendel2020semi}  & - & - & - & 73.0 & 74.7 & - \\
SN\cite{zhang2020robust1} & - & - & - & - & - & \textbf{76.7} \\
Alonso et al.\cite{alonso2021semi} & - & \textbf{67.9} & 70.0 & 71.6 & - & 74.1 \\
\hline
Ours        & 56.8 & 66.8 & \textbf{72.4} & \textbf{73.4} & \textbf{74.9} & 76.6 \\
\emph{Rela. Impr.} ($\%$) & \emph{33.3} & \emph{25.8} & \emph{20.5} & \emph{12.6} & \emph{8.7} & \emph{4.4} \\
\bottomrule
\end{tabular}
\label{table:pascal}
\end{table}

\begin{table}[h]
\centering
\caption{Comparison between our proposed method and SOTA semi-supervised methods on Cityscapes, where DeepLab-V2 is a supervised baseline. $^*$: DeepLab-v2 backbone and DeepLab-v3 corrector for ECS.}
\begin{tabular}{m{2.8cm}<{\centering}m{0.6cm}<{\centering}m{0.6cm}<{\centering}m{0.6cm}<{\centering}m{0.6cm}<{\centering}m{0.6cm}<{\centering}}
\toprule
Methods    &1/30 &1/8 & 1/4 &1/2 &full\\
\hline
DeepLab-v2\cite{chen2017deeplab}  & 43.4 & 55.0 & 60.3 & 62.2 & 66.2 \\
\hline
AdvSemiSeg\cite{hung2018adversarial}  & -    & 58.8 & 62.3 & 65.7 & 67.7 \\
CowMix\cite{french2019semi}      & 49.0 & 60.5 & 64.1 & 66.5 & 69.0 \\
MLMT+S4GAN\cite{mittal2019semi}  & -    & 59.3 & 61.9 & -    & 65.8 \\
DST-CBC\cite{feng2020semi}     & 48.7 & 60.5 & 64.4 & -    & 66.9 \\
ClassMix\cite{olsson2020classmix}    & 54.1 & 61.4 & 63.6 & 66.3 & -    \\
ECS$^*$ \cite{mendel2020semi} & - & 60.3 & 63.8 & - & - \\
\emph{C}$^3$-SemiSeg\cite{zhou2021c3}  & 55.2 & \textbf{63.2} & 65.5 & - & \textbf{69.5} \\
\hline
Ours        & \textbf{56.3} & 62.3 & \textbf{66.2} & \textbf{67.4} & 69.0 \\
\emph{Rela. Impr.} ($\%$) & \emph{29.7} & \emph{13.3} & \emph{9.8} & \emph{8.4} & \emph{4.2}  \\
\bottomrule
\end{tabular}
\label{table:city}
\end{table}

\begin{table}[h]
\centering
\caption{Comparison between our proposed method and SOTA semi-supervised methods on PASCAL-Context, where DeepLab-V2 is a supervised baseline.}
\begin{tabular}{m{3cm}<{\centering}m{0.6cm}<{\centering}m{0.6cm}<{\centering}m{0.6cm}<{\centering}}
\toprule
Methods    &1/8 & 1/4 &full\\
\hline
DeepLab-v2\cite{chen2017deeplab}   & 32.1 & 35.4 & 41.0 \\
\hline
AdvSemiSeg\cite{hung2018adversarial}   & 32.8 & 34.8 & 39.1 \\
MLMT+S4GAN\cite{mittal2019semi}    & 35.3 & 37.8 & 41.1 \\
\hline
Ours         & \textbf{36.1} & \textbf{39.1} & \textbf{42.5} \\
\emph{Rela. Impr.} ($\%$)  & \emph{12.5} & \emph{10.5} & \emph{3.7}  \\
\bottomrule
\end{tabular}
\label{table:pc}
\end{table}

\begin{table}[h]
\centering
\caption{Comparison between our proposed method with SOTA semi-supervised methods on PASCAL VOC2012, where DeepLab-V3+ is a supervised baseline. Note that the same baseline achieves different segmentation results in different works due to different implemenation details.}
\begin{tabular}{m{2.5cm}<{\centering}m{1.5cm}<{\centering}m{1cm}<{\centering}m{1cm}<{\centering}}
\toprule
Methods     &1/8 & 1/4 \\
\hline
Baseline                                 & 65.2 & 69.8 \\
AdvSemiSeg\cite{hung2018adversarial}     & 66.2 & 70.5  \\
\emph{Rela. Impr.} ($\%$)                & \emph{1.5} & \emph{1.0}  \\
\hline
Baseline                                 & 68.1 & 72.4 \\
CCT\cite{ouali2020semi}                  & 70.9 & 73.4  \\
\emph{Rela. Impr.} ($\%$)                & \emph{4.1} & \emph{1.4}  \\
\hline
Baseline                                 & 65.2 & 69.8 \\
ECS\cite{mendel2020semi}                 & 70.2 & 72.6  \\
\emph{Rela. Impr.} ($\%$)                & \emph{7.7} & \emph{4.0}  \\
\hline
Baseline                                 & 68.1 & 72.4 \\
CPS\cite{chen2021semi}                   & 73.7 & 74.9  \\
\emph{Rela. Impr.} ($\%$)                & \emph{8.2} & \emph{3.5}  \\
\hline
Baseline                                 & 68.3 & 70.5 \\
ST\cite{yang2022st++}                    & 73.3 & 75.0  \\
\emph{Rela. Impr.} ($\%$)                & \emph{7.3} & \textbf{\emph{6.4}}  \\
\hline
Baseline                                 & 65.2 & 69.6 \\
Ours                                     & 70.8 & 72.8  \\
\emph{Rela. Impr.} ($\%$)                & \textbf{\emph{8.6}} & \emph{4.6}  \\
\bottomrule
\end{tabular}
\label{table:pascal2}
\end{table}

\subsubsection{Results on PASCAL VOC2012}
%
Following previous works, we conduct experiments on eight proportions of labeled data and results are summarized in Table \ref{table:pascal}.
Our results are compared with seven recent state-of-the-art works, all of which use the same DeepLab-v2 network.
Our method outperforms the supervised baselines, i.e. DeppLab-v2, by a large margin. Besides, the less labeled data, the higher relative improvement. Specifically, when only 1/100 of labeled data is available, our solution improves relatively over baseline by up to
33.3\% and with 1/8 training data, we can achieve the same performance of 73.4\% in mIoU as fully-supervised DeepLab-v2.
These results also demonstrate that our proposed method achieves the best results on PASCAL VOC2012, suggesting that our proposed geometry-based consistency regularization and Adversarial Dual-Student framework can both be successfully applied to semi-supervised semantic segmentation.
One reason for the higher performance is that our proposed DGW mechanism introduces complex and flexible geometric perturbation. It enriches diversities in consistency regularization and meanwhile avoids severe context destruction.
The other one is that Adversarial Dual-Student can avoid the limitation of Mean-Teacher framework and better reveal the potential of unlabeled samples.

The qualitative comparison results are demonstrated in Fig. \ref{fig:vs_pascal}. More visual comparison results are provided in the supplementary file.
These is a clear improvement over the baseline and other state-of-the-art methods.
Notice that one significant advantage of our method is that the segmentation results are more complete and have less wrong labels. 

Besides DeepLab-v2 as the baseline network, we also conduct experiments with DeepLab-v3+ as the baseline. The results are summarized in Table \ref{table:pascal2}. Note that, the performance of the baseline network (i.e., DeepLab-v3+) is different in \cite{mendel2020semi,fan2022ucc}, and \cite{yang2022st++} due to different implementation details (such as hyperparameters, pre-processing). Therefore, it is unfair to directly compare the final performance for semi-supervised learning with different baseline realizations. Hence, we compare the relative improvements over baselines for different semi-supervised learning methods. When only 1/8 of labeled data is available, our solution improves relatively over baseline by up to 8.6\%, which outperforms other state-of-the-art works. When only 1/4 of labeled data is available, our solution achieves the second-best results.

\subsubsection{Results on Cityscapes and PASCAL-Context}
Table \ref{table:city} presents the results on Cityscapes under settings of different labeled data volumn.
These results demonstrate that our proposed method also significantly and consistently outperforms the baseline model and state-of-the-art semi-supervised methods on Cityscapes. Especially, our proposed method with only 1/4 labeled data achieves the same performance as fully-supervised DeepLab-v2.

%
The qualitative results with 1/4 labeled samples are demonstrated in Fig \ref{fig:vs_city}. Since the differences on the Cityscapes are subtle, in each segmentation map, we add the zoomed-in views of informative areas.

Table \ref{table:pc} presents the results on PASCAL-Context dataset with different amounts of labeled data. PASCAL-Context dataset is smaller and more difficult than PASCAL VOC. For this dataset, our proposed method still achieves better performance than the baseline model and state-of-the-art semi-supervised methods. Specifically, when 1/8, 1/4, and full labeled data are used, our solution improves relatively over baseline by 12.5\%, 10.5\%, and 3.7\% respectively.

\section{Conclusion}
In this work, a novel ADS framework with DGW is proposed specially for semi-supervised semantic segmentation task. The proposed DGW is a context-friendly unsupervised data augmentation strategy for unlabeled image perturbation and it is verified to be more effective than existing mechanisms to construct consistency regularization. The ADS framework is also validated to be better than the popular Mean-Teacher framework and state-of-the-art performance can be achieved on both PASCAL VOC2012 and Cityscapes datasets.

Recently, self-paced learning (SPL) \cite{kumar2010self,zhang2018spftn,zhang2019leveraging,zhang2020few} is utilized to better exploit the reliable and unreliable labels from the pseudo labels. Among them, adversarial-paced learning (APL) \cite{zhang2020few} uses adversarial learning to improve the performance and flexibility of self-paced learning. In the future, we would like to improve our adversarial learning based on SPL.

%

%

\ifCLASSOPTIONcaptionsoff
  \newpage
\fi



\bibliographystyle{IEEEtran}
\bibliography{egbib}

\begin{thebibliography}{10}
\providecommand{\url}[1]{#1}
\csname url@samestyle\endcsname
\providecommand{\newblock}{\relax}
\providecommand{\bibinfo}[2]{#2}
\providecommand{\BIBentrySTDinterwordspacing}{\spaceskip=0pt\relax}
\providecommand{\BIBentryALTinterwordstretchfactor}{4}
\providecommand{\BIBentryALTinterwordspacing}{\spaceskip=\fontdimen2\font plus
\BIBentryALTinterwordstretchfactor\fontdimen3\font minus
  \fontdimen4\font\relax}
\providecommand{\BIBforeignlanguage}[2]{{%
\expandafter\ifx\csname l@#1\endcsname\relax
\typeout{** WARNING: IEEEtran.bst: No hyphenation pattern has been}%
\typeout{** loaded for the language `#1'. Using the pattern for}%
\typeout{** the default language instead.}%
\else
\language=\csname l@#1\endcsname
\fi
#2}}
\providecommand{\BIBdecl}{\relax}
\BIBdecl

\bibitem{chen2014semantic}
L.-C. Chen, G.~Papandreou, I.~Kokkinos, K.~Murphy, and A.~L. Yuille, ``Semantic
  image segmentation with deep convolutional nets and fully connected crfs,''
  \emph{arXiv preprint arXiv:1412.7062}, 2014.

\bibitem{ronneberger2015u}
O.~Ronneberger, P.~Fischer, and T.~Brox, ``U-net: Convolutional networks for
  biomedical image segmentation,'' in \emph{International Conference on Medical
  image computing and computer-assisted intervention}.\hskip 1em plus 0.5em
  minus 0.4em\relax Springer, 2015, pp. 234--241.

\bibitem{chen2017deeplab}
L.-C. Chen, G.~Papandreou, I.~Kokkinos, K.~Murphy, and A.~L. Yuille, ``Deeplab:
  Semantic image segmentation with deep convolutional nets, atrous convolution,
  and fully connected crfs,'' \emph{IEEE transactions on pattern analysis and
  machine intelligence}, vol.~40, no.~4, pp. 834--848, 2017.

\bibitem{chen2017rethinking}
L.-C. Chen, G.~Papandreou, F.~Schroff, and H.~Adam, ``Rethinking atrous
  convolution for semantic image segmentation,'' \emph{arXiv preprint
  arXiv:1706.05587}, 2017.

\bibitem{zhao2017pyramid}
H.~Zhao, J.~Shi, X.~Qi, X.~Wang, and J.~Jia, ``Pyramid scene parsing network,''
  in \emph{Proceedings of the IEEE conference on computer vision and pattern
  recognition}, 2017, pp. 2881--2890.

\bibitem{yuan2018ocnet}
Y.~Yuan and J.~Wang, ``Ocnet: Object context network for scene parsing,''
  \emph{arXiv preprint arXiv:1809.00916}, 2018.

\bibitem{sun2019high}
K.~Sun, Y.~Zhao, B.~Jiang, T.~Cheng, B.~Xiao, D.~Liu, Y.~Mu, X.~Wang, W.~Liu,
  and J.~Wang, ``High-resolution representations for labeling pixels and
  regions,'' \emph{arXiv preprint arXiv:1904.04514}, 2019.

\bibitem{ji2020encoder}
J.~Ji, R.~Shi, S.~Li, P.~Chen, and Q.~Miao, ``Encoder-decoder with cascaded
  crfs for semantic segmentation,'' \emph{IEEE Transactions on Circuits and
  Systems for Video Technology}, 2020.

\bibitem{athanasiadis2007semantic}
T.~Athanasiadis, P.~Mylonas, Y.~Avrithis, and S.~Kollias, ``Semantic image
  segmentation and object labeling,'' \emph{IEEE transactions on circuits and
  systems for video technology}, vol.~17, no.~3, pp. 298--312, 2007.

\bibitem{meng2019weakly}
F.~Meng, K.~Luo, H.~Li, Q.~Wu, and X.~Xu, ``Weakly supervised semantic
  segmentation by a class-level multiple group cosegmentation and foreground
  fusion strategy,'' \emph{IEEE Transactions on Circuits and Systems for Video
  Technology}, vol.~30, no.~12, pp. 4823--4836, 2019.

\bibitem{yun2019cutmix}
S.~Yun, D.~Han, S.~J. Oh, S.~Chun, J.~Choe, and Y.~Yoo, ``Cutmix:
  Regularization strategy to train strong classifiers with localizable
  features,'' in \emph{Proceedings of the IEEE International Conference on
  Computer Vision}, 2019, pp. 6023--6032.

\bibitem{french2019semi}
G.~French, T.~Aila, S.~Laine, M.~Mackiewicz, and G.~Finlayson,
  ``Semi-supervised semantic segmentation needs strong, high-dimensional
  perturbations,'' \emph{arXiv preprint arXiv:1906.01916}, 2019.

\bibitem{olsson2020classmix}
V.~Olsson, W.~Tranheden, J.~Pinto, and L.~Svensson, ``Classmix:
  Segmentation-based data augmentation for semi-supervised learning,''
  \emph{arXiv preprint arXiv:2007.07936}, 2020.

\bibitem{tarvainen2017mean}
A.~Tarvainen and H.~Valpola, ``Mean teachers are better role models:
  Weight-averaged consistency targets improve semi-supervised deep learning
  results,'' in \emph{Advances in Neural Information Processing Systems
  (NeurIPS)}, 2017, pp. 1195--1204.

\bibitem{ke2019dual}
Z.~Ke, D.~Wang, Q.~Yan, J.~Ren, and R.~W. Lau, ``Dual student: Breaking the
  limits of the teacher in semi-supervised learning,'' in \emph{Proceedings of
  the IEEE International Conference on Computer Vision}, 2019, pp. 6728--6736.

\bibitem{bookstein1989principal}
F.~L. Bookstein, ``Principal warps: Thin-plate splines and the decomposition of
  deformations,'' \emph{IEEE Transactions on pattern analysis and machine
  intelligence}, vol.~11, no.~6, pp. 567--585, 1989.

\bibitem{laine2016temporal}
S.~Laine and T.~Aila, ``Temporal ensembling for semi-supervised learning,''
  \emph{arXiv preprint arXiv:1610.02242}, 2016.

\bibitem{miyato2018virtual}
T.~Miyato, S.-i. Maeda, M.~Koyama, and S.~Ishii, ``Virtual adversarial
  training: a regularization method for supervised and semi-supervised
  learning,'' \emph{IEEE Transactions on Pattern Analysis and Machine
  Intelligence (TPAMI)}, vol.~41, no.~8, pp. 1979--1993, 2018.

\bibitem{ji2019invariant}
X.~Ji, J.~F. Henriques, and A.~Vedaldi, ``Invariant information clustering for
  unsupervised image classification and segmentation,'' in \emph{Proceedings of
  the IEEE International Conference on Computer Vision (ICCV)}, 2019, pp.
  9865--9874.

\bibitem{bachman2019learning}
P.~Bachman, R.~D. Hjelm, and W.~Buchwalter, ``Learning representations by
  maximizing mutual information across views,'' in \emph{Advances in Neural
  Information Processing Systems (NeurIPS)}, 2019, pp. 15\,509--15\,519.

\bibitem{berthelot2019mixmatch}
D.~Berthelot, N.~Carlini, I.~Goodfellow, N.~Papernot, A.~Oliver, and C.~A.
  Raffel, ``Mixmatch: A holistic approach to semi-supervised learning,'' in
  \emph{Advances in Neural Information Processing Systems (NeurIPS)}, 2019, pp.
  5050--5060.

\bibitem{lee2013pseudo}
D.-H. Lee, ``Pseudo-label: The simple and efficient semi-supervised learning
  method for deep neural networks,'' in \emph{International Conference on
  Machine Learning (ICML) Workshop}, vol.~3, 2013, p.~2.

\bibitem{ouali2020semi}
Y.~Ouali, C.~Hudelot, and M.~Tami, ``Semi-supervised semantic segmentation with
  cross-consistency training,'' in \emph{Proceedings of the IEEE/CVF Conference
  on Computer Vision and Pattern Recognition}, 2020, pp. 12\,674--12\,684.

\bibitem{chen2021semi}
X.~Chen, Y.~Yuan, G.~Zeng, and J.~Wang, ``Semi-supervised semantic segmentation
  with cross pseudo supervision,'' in \emph{Proceedings of the IEEE/CVF
  Conference on Computer Vision and Pattern Recognition}, 2021, pp. 2613--2622.

\bibitem{liu2022perturbed}
Y.~Liu, Y.~Tian, Y.~Chen, F.~Liu, V.~Belagiannis, and G.~Carneiro, ``Perturbed
  and strict mean teachers for semi-supervised semantic segmentation,'' in
  \emph{Proceedings of the IEEE/CVF Conference on Computer Vision and Pattern
  Recognition}, 2022, pp. 4258--4267.

\bibitem{souly2017semi}
N.~Souly, C.~Spampinato, and M.~Shah, ``Semi supervised semantic segmentation
  using generative adversarial network,'' in \emph{Proceedings of the IEEE
  International Conference on Computer Vision}, 2017, pp. 5688--5696.

\bibitem{hung2018adversarial}
W.-C. Hung, Y.-H. Tsai, Y.-T. Liou, Y.-Y. Lin, and M.-H. Yang, ``Adversarial
  learning for semi-supervised semantic segmentation,'' \emph{arXiv preprint
  arXiv:1802.07934}, 2018.

\bibitem{feng2020semi}
Z.~Feng, Q.~Zhou, G.~Cheng, X.~Tan, J.~Shi, and L.~Ma, ``Semi-supervised
  semantic segmentation via dynamic self-training and class-balanced
  curriculum,'' \emph{arXiv preprint arXiv:2004.08514}, 2020.

\bibitem{yang2022st++}
L.~Yang, W.~Zhuo, L.~Qi, Y.~Shi, and Y.~Gao, ``St++: Make self-training work
  better for semi-supervised semantic segmentation,'' in \emph{Proceedings of
  the IEEE/CVF Conference on Computer Vision and Pattern Recognition}, 2022,
  pp. 4268--4277.

\bibitem{fan2022ucc}
J.~Fan, B.~Gao, H.~Jin, and L.~Jiang, ``Ucc: Uncertainty guided cross-head
  co-training for semi-supervised semantic segmentation,'' in \emph{Proceedings
  of the IEEE/CVF Conference on Computer Vision and Pattern Recognition}, 2022,
  pp. 9947--9956.

\bibitem{mendel2020semi}
R.~Mendel, L.~A. de~Souza, D.~Rauber, J.~P. Papa, and C.~Palm,
  ``Semi-supervised segmentation based on error-correcting supervision,'' in
  \emph{European Conference on Computer Vision}.\hskip 1em plus 0.5em minus
  0.4em\relax Springer, 2020, pp. 141--157.

\bibitem{zhang2020robust}
J.~Zhang, Z.~Li, C.~Zhang, and H.~Ma, ``Robust adversarial learning for
  semi-supervised semantic segmentation,'' in \emph{2020 IEEE International
  Conference on Image Processing (ICIP)}.\hskip 1em plus 0.5em minus
  0.4em\relax IEEE, 2020, pp. 728--732.

\bibitem{zhang2020robust1}
------, ``Robust semi-supervised semantic segmentation based on self-attention
  and spectral normalization,'' in \emph{2020 International Joint Conference on
  Neural Networks (IJCNN)}.\hskip 1em plus 0.5em minus 0.4em\relax IEEE, 2020,
  pp. 1--8.

\bibitem{chen2020digging}
Z.~Chen, R.~Zhang, G.~Zhang, Z.~Ma, and T.~Lei, ``Digging into pseudo label: a
  low-budget approach for semi-supervised semantic segmentation,'' \emph{IEEE
  Access}, vol.~8, pp. 41\,830--41\,837, 2020.

\bibitem{alonso2021semi}
I.~Alonso, A.~Sabater, D.~Ferstl, L.~Montesano, and A.~C. Murillo,
  ``Semi-supervised semantic segmentation with pixel-level contrastive learning
  from a class-wise memory bank,'' in \emph{Proceedings of the IEEE/CVF
  International Conference on Computer Vision}, 2021, pp. 8219--8228.

\bibitem{zhou2021c3}
Y.~Zhou, H.~Xu, W.~Zhang, B.~Gao, and P.-A. Heng, ``C3-semiseg: Contrastive
  semi-supervised segmentation via cross-set learning and dynamic
  class-balancing,'' in \emph{Proceedings of the IEEE/CVF International
  Conference on Computer Vision}, 2021, pp. 7036--7045.

\bibitem{papandreou2015weakly}
G.~Papandreou, L.-C. Chen, K.~P. Murphy, and A.~L. Yuille, ``Weakly-and
  semi-supervised learning of a deep convolutional network for semantic image
  segmentation,'' in \emph{Proceedings of the IEEE international conference on
  computer vision}, 2015, pp. 1742--1750.

\bibitem{li2018weakly}
Q.~Li, A.~Arnab, and P.~H. Torr, ``Weakly-and semi-supervised panoptic
  segmentation,'' in \emph{Proceedings of the European Conference on Computer
  Vision (ECCV)}, 2018, pp. 102--118.

\bibitem{lee2019ficklenet}
J.~Lee, E.~Kim, S.~Lee, J.~Lee, and S.~Yoon, ``Ficklenet: Weakly and
  semi-supervised semantic image segmentation using stochastic inference,'' in
  \emph{Proceedings of the IEEE conference on computer vision and pattern
  recognition}, 2019, pp. 5267--5276.

\bibitem{ibrahim2020semi}
M.~S. Ibrahim, A.~Vahdat, M.~Ranjbar, and W.~G. Macready, ``Semi-supervised
  semantic image segmentation with self-correcting networks,'' in
  \emph{Proceedings of the IEEE/CVF Conference on Computer Vision and Pattern
  Recognition}, 2020, pp. 12\,715--12\,725.

\bibitem{kalluri2019universal}
T.~Kalluri, G.~Varma, M.~Chandraker, and C.~Jawahar, ``Universal
  semi-supervised semantic segmentation,'' in \emph{Proceedings of the IEEE
  International Conference on Computer Vision}, 2019, pp. 5259--5270.

\bibitem{sedai2017semi}
S.~Sedai, D.~Mahapatra, S.~Hewavitharanage, S.~Maetschke, and R.~Garnavi,
  ``Semi-supervised segmentation of optic cup in retinal fundus images using
  variational autoencoder,'' in \emph{International Conference on Medical Image
  Computing and Computer-Assisted Intervention}.\hskip 1em plus 0.5em minus
  0.4em\relax Springer, 2017, pp. 75--82.

\bibitem{chartsias2018factorised}
A.~Chartsias, T.~Joyce, G.~Papanastasiou, S.~Semple, M.~Williams, D.~Newby,
  R.~Dharmakumar, and S.~A. Tsaftaris, ``Factorised spatial representation
  learning: Application in semi-supervised myocardial segmentation,'' in
  \emph{International Conference on Medical Image Computing and
  Computer-Assisted Intervention}.\hskip 1em plus 0.5em minus 0.4em\relax
  Springer, 2018, pp. 490--498.

\bibitem{cui2019semi}
W.~Cui, Y.~Liu, Y.~Li, M.~Guo, Y.~Li, X.~Li, T.~Wang, X.~Zeng, and C.~Ye,
  ``Semi-supervised brain lesion segmentation with an adapted mean teacher
  model,'' in \emph{International Conference on Information Processing in
  Medical Imaging}.\hskip 1em plus 0.5em minus 0.4em\relax Springer, 2019, pp.
  554--565.

\bibitem{li2020self}
Y.~Li, J.~Chen, X.~Xie, K.~Ma, and Y.~Zheng, ``Self-loop uncertainty: A novel
  pseudo-label for semi-supervised medical image segmentation,'' in
  \emph{International Conference on Medical Image Computing and
  Computer-Assisted Intervention}.\hskip 1em plus 0.5em minus 0.4em\relax
  Springer, 2020, pp. 614--623.

\bibitem{hang2020local}
W.~Hang, W.~Feng, S.~Liang, L.~Yu, Q.~Wang, K.-S. Choi, and J.~Qin, ``Local and
  global structure-aware entropy regularized mean teacher model for 3d left
  atrium segmentation,'' in \emph{International Conference on Medical Image
  Computing and Computer-Assisted Intervention}.\hskip 1em plus 0.5em minus
  0.4em\relax Springer, 2020, pp. 562--571.

\bibitem{mittal2019semi}
S.~Mittal, M.~Tatarchenko, and T.~Brox, ``Semi-supervised semantic segmentation
  with high-and low-level consistency,'' \emph{IEEE Transactions on Pattern
  Analysis and Machine Intelligence}, 2019.

\bibitem{zhang2017mixup}
H.~Zhang, M.~Cisse, Y.~N. Dauphin, and D.~Lopez-Paz, ``mixup: Beyond empirical
  risk minimization,'' \emph{arXiv preprint arXiv:1710.09412}, 2017.

\bibitem{cubuk2018autoaugment}
E.~D. Cubuk, B.~Zoph, D.~Mane, V.~Vasudevan, and Q.~V. Le, ``Autoaugment:
  Learning augmentation policies from data,'' \emph{arXiv preprint
  arXiv:1805.09501}, 2018.

\bibitem{ict}
V.~Verma, A.~Lamb, J.~Kannala, Y.~Bengio, and D.~Lopez-Paz, ``Interpolation
  consistency training for semi-supervised learning,'' \emph{arXiv preprint
  arXiv:1903.03825}, 2019.

\bibitem{zhao2019data}
A.~Zhao, G.~Balakrishnan, F.~Durand, J.~V. Guttag, and A.~V. Dalca, ``Data
  augmentation using learned transformations for one-shot medical image
  segmentation,'' in \emph{Proceedings of the IEEE conference on computer
  vision and pattern recognition}, 2019, pp. 8543--8553.

\bibitem{lee2020learning}
D.~Lee, H.~Park, T.~Pham, and C.~D. Yoo, ``Learning augmentation network via
  influence functions,'' in \emph{Proceedings of the IEEE/CVF Conference on
  Computer Vision and Pattern Recognition}, 2020, pp. 10\,961--10\,970.

\bibitem{wang2020regularizing}
Y.~Wang, G.~Huang, S.~Song, X.~Pan, Y.~Xia, and C.~Wu, ``Regularizing deep
  networks with semantic data augmentation,'' \emph{arXiv preprint
  arXiv:2007.10538}, 2020.

\bibitem{jakab2018unsupervised}
T.~Jakab, A.~Gupta, H.~Bilen, and A.~Vedaldi, ``Unsupervised learning of object
  landmarks through conditional image generation,'' in \emph{Advances in neural
  information processing systems}, 2018, pp. 4016--4027.

\bibitem{zhang2018unsupervised}
Y.~Zhang, Y.~Guo, Y.~Jin, Y.~Luo, Z.~He, and H.~Lee, ``Unsupervised discovery
  of object landmarks as structural representations,'' in \emph{Proceedings of
  the IEEE Conference on Computer Vision and Pattern Recognition}, 2018, pp.
  2694--2703.

\bibitem{everingham2015pascal}
M.~Everingham, S.~A. Eslami, L.~Van~Gool, C.~K. Williams, J.~Winn, and
  A.~Zisserman, ``The pascal visual object classes challenge: A
  retrospective,'' \emph{International journal of computer vision}, vol. 111,
  no.~1, pp. 98--136, 2015.

\bibitem{hariharan2011semantic}
B.~Hariharan, P.~Arbel{\'a}ez, L.~Bourdev, S.~Maji, and J.~Malik, ``Semantic
  contours from inverse detectors,'' in \emph{2011 International Conference on
  Computer Vision}.\hskip 1em plus 0.5em minus 0.4em\relax IEEE, 2011, pp.
  991--998.

\bibitem{cordts2016cityscapes}
M.~Cordts, M.~Omran, S.~Ramos, T.~Rehfeld, M.~Enzweiler, R.~Benenson,
  U.~Franke, S.~Roth, and B.~Schiele, ``The cityscapes dataset for semantic
  urban scene understanding,'' in \emph{Proceedings of the IEEE conference on
  computer vision and pattern recognition}, 2016, pp. 3213--3223.

\bibitem{mottaghi2014role}
R.~Mottaghi, X.~Chen, X.~Liu, N.-G. Cho, S.-W. Lee, S.~Fidler, R.~Urtasun, and
  A.~Yuille, ``The role of context for object detection and semantic
  segmentation in the wild,'' in \emph{Proceedings of the IEEE conference on
  computer vision and pattern recognition}, 2014, pp. 891--898.

\bibitem{He_2016_CVPR}
K.~He, X.~Zhang, S.~Ren, and J.~Sun, ``Deep residual learning for image
  recognition,'' in \emph{Proceedings of the IEEE Conference on Computer Vision
  and Pattern Recognition}, 2016, pp. 770--778.

\bibitem{lin2014microsoft}
T.-Y. Lin, M.~Maire, S.~Belongie, J.~Hays, P.~Perona, D.~Ramanan,
  P.~Doll{\'a}r, and C.~L. Zitnick, ``Microsoft coco: Common objects in
  context,'' in \emph{European conference on computer vision}.\hskip 1em plus
  0.5em minus 0.4em\relax Springer, 2014, pp. 740--755.

\bibitem{kumar2010self}
M.~P. Kumar, B.~Packer, and D.~Koller, ``Self-paced learning for latent
  variable models,'' in \emph{Advances in Neural Information Processing Systems
  (NeurIps)}, 2010, pp. 1189--1197.

\bibitem{zhang2018spftn}
D.~Zhang, J.~Han, L.~Yang, and D.~Xu, ``Spftn: A joint learning framework for
  localizing and segmenting objects in weakly labeled videos,'' \emph{IEEE
  transactions on pattern analysis and machine intelligence}, vol.~42, no.~2,
  pp. 475--489, 2018.

\bibitem{zhang2019leveraging}
D.~Zhang, J.~Han, L.~Zhao, and D.~Meng, ``Leveraging prior-knowledge for weakly
  supervised object detection under a collaborative self-paced curriculum
  learning framework,'' \emph{International Journal of Computer Vision}, vol.
  127, no.~4, pp. 363--380, 2019.

\bibitem{zhang2020few}
D.~Zhang, H.~Tian, and J.~Han, ``Few-cost salient object detection with
  adversarial-paced learning,'' \emph{Advances in Neural Information Processing
  Systems}, vol.~33, pp. 12\,236--12\,247, 2020.

\end{thebibliography}
\end{document}